\documentclass[letterpaper, 10 pt, journal, twoside]{IEEEtran}

\usepackage{amsmath,amsfonts}
\usepackage{algpseudocode} 
\usepackage{algorithm}
\usepackage{array}
\usepackage[caption=false,font=footnotesize,labelfont=rm,textfont=rm]{subfig}

\usepackage{textcomp}
\usepackage{stfloats}
\usepackage{url}
\usepackage{verbatim}
\usepackage{graphicx}
\usepackage{cite}
\usepackage{bbding}
\usepackage{mathtools}
\begin{document}
\title{Guided Time-optimal Model Predictive Control of a Multi-rotor}

\author{Guangyu Zhang, Yongjie Zheng, Yuqing He, Liying Yang, Hongyu Nie, Chaoxiong Huang, Yiwen Zhao
\thanks{This work was supported in part by the National Natural Science Foundation of China under Grant 61991413 and 91948303, National Natural Science Foundation of China Innovation Research Group Project under Grant 61821005, Youth Innovation Promotion Association under Grant Y2022065, and Shenyang Science and Technology Plan under Grant 21-108-9-18.}
\thanks{
    G. Zhang, Y. He, L. Yang, C. Huang,  and Y. Zhao are with the State Key Laboratory of Robotics, Shenyang Institute of Automation, Chinese Academy of
    Sciences, Shenyang 110016, China, and also with the Institute for Robotics and Intelligent Manufacturing, Chinese Academy of
    Sciences, Shenyang 110016, China(e-mail: zhangguangyu@sia.cn).
  }
  \thanks{
    Y. Zheng and H. Nie are with the State Key Laboratory of Robotics, Shenyang Institute
    of Automation, Chinese Academy of Sciences, Shenyang 110016, China, with
    the Institutes for Robotics and Intelligent Manufacturing, Chinese Academy of
    Sciences, Shenyang 110016, China, and also with the University of Chinese
    Academy of Sciences, Beijing 100049, China(e-mail: zhengyongjie@sia.cn).
  }}
\markboth{IEEE Control Systems Letters (Accepted Version)}%
{Shell \MakeLowercase{\textit{et al.}}: A Sample Article Using IEEEtran.cls for IEEE Journals}


\maketitle

\begin{abstract}
  Time-optimal control of a multi-rotor remains an open problem due to the under-actuation and nonlinearity of
  its dynamics, which make it difficult to solve this problem directly.  In this paper, the time-optimal control problem of the multi-rotor
is studied.  Firstly, a thrust limit optimal decomposition method is proposed, 
  which can reasonably decompose the limited thrust into three directions according to the current state and the target state. 
  As a result, the thrust limit constraint is decomposed as a linear constraint. 
  With the linear constraint and decoupled dynamics, a time-optimal guidance trajectory can be obtained. 
  Then, a cost function is defined based on the time-optimal guidance trajectory, which has a quadratic form and can be used to evaluate the time-optimal performance of the system outputs. 
  Finally, based on the cost function, the time-optimal control problem is reformulated as an MPC (Model Predictive Control) problem. The experimental results demonstrate the feasibility and validity of the proposed methods. 
\end{abstract}


\section{Introduction}
\IEEEPARstart{T}{ime-optimal} flight control of a multi-rotor is a problem considering controlling the multi-rotor flight from an initial state to a terminal state in the minimum time. The maneuver performance of the multi-rotor can be maximized with a time-optimal controller, which is crucial for some applications , such as dynamic target catching and drone racing \cite{zhang_grasp_2018, foehn2021time, zhang2019optimal}. 
To improve the time-optimal performance of the multi-rotor, trajectory optimization and optimal control strategies have been used in the existing literature.

For the studies using trajectory optimization strategy, to ensure smoothness and dynamic feasibility, the time-parameterized high-order polynomial trajectory optimization schemes based on the system's differential flatness property have been proposed.
The differential flatness property implies that the system's state and inputs can be computed by its flat outputs and their high-order derivatives\cite{mellinger2011minimum}. Therefore,  by utilizing a high-order polynomial trajectory of the flat outputs of the multi-rotor,  i.e. position and yaw angle,  the smoothness of its state can be ensured.
In the works of \cite{van2013time} and \cite{spedicato2017minimum}, the optimization of polynomial trajectories is achieved by formulating a fixed end-time-optimal problem using a variable associated with the path curve. By solving this problem, piecewise trajectories that exhibit the minimum time property can be obtained.
Similarly, in \cite{gao2018optimal}, a similar approach is proposed where time is allocated along a path with fixed geometry to generate a minimum time trajectory.
Other studies have proposed methods based on Sequential Convex Programming (SCP), which involve allocating time into a sequence of discrete-time waypoints while considering dynamic constraints \cite{hu2017time}, \cite{8206522}.
However, when the number of waypoints is large, the optimization of these waypoints can become time-consuming. Consequently, these methods are better suited for offline trajectory planning tasks.

For the studies using optimal control strategy, some simplified dynamic models have been used to solve the time-optimal control problem. 
In \cite{hehn2012real,hehn2012performance},  a time-optimal control problem is formulated  based on a simplified two-dimensional dynamic model, in which the thrust and rotational rate are used
as control inputs, and a bang-off-bang controller is designed using Pontryagin's Minimum Principle.
Point-mass dynamics is a simple linear dynamic model, so, the analytical form solution of the minimum time control problem of the point-mass system can be obtained using Pontryagin's Minimum Principle \cite{he2020time}, \cite{haschke2008line}. 
Building upon this analytical  solution, a minimum time motion primitive approach is proposed in \cite{liu2017search}, \cite{penicka2022minimum}, and a real-time time-optimal trajectory generation approach is proposed in \cite{beul2016analytical,hehn2015real}. 
\begin{figure}[ht]
  \centering
  \subfloat[]{\includegraphics[width=0.45\linewidth]{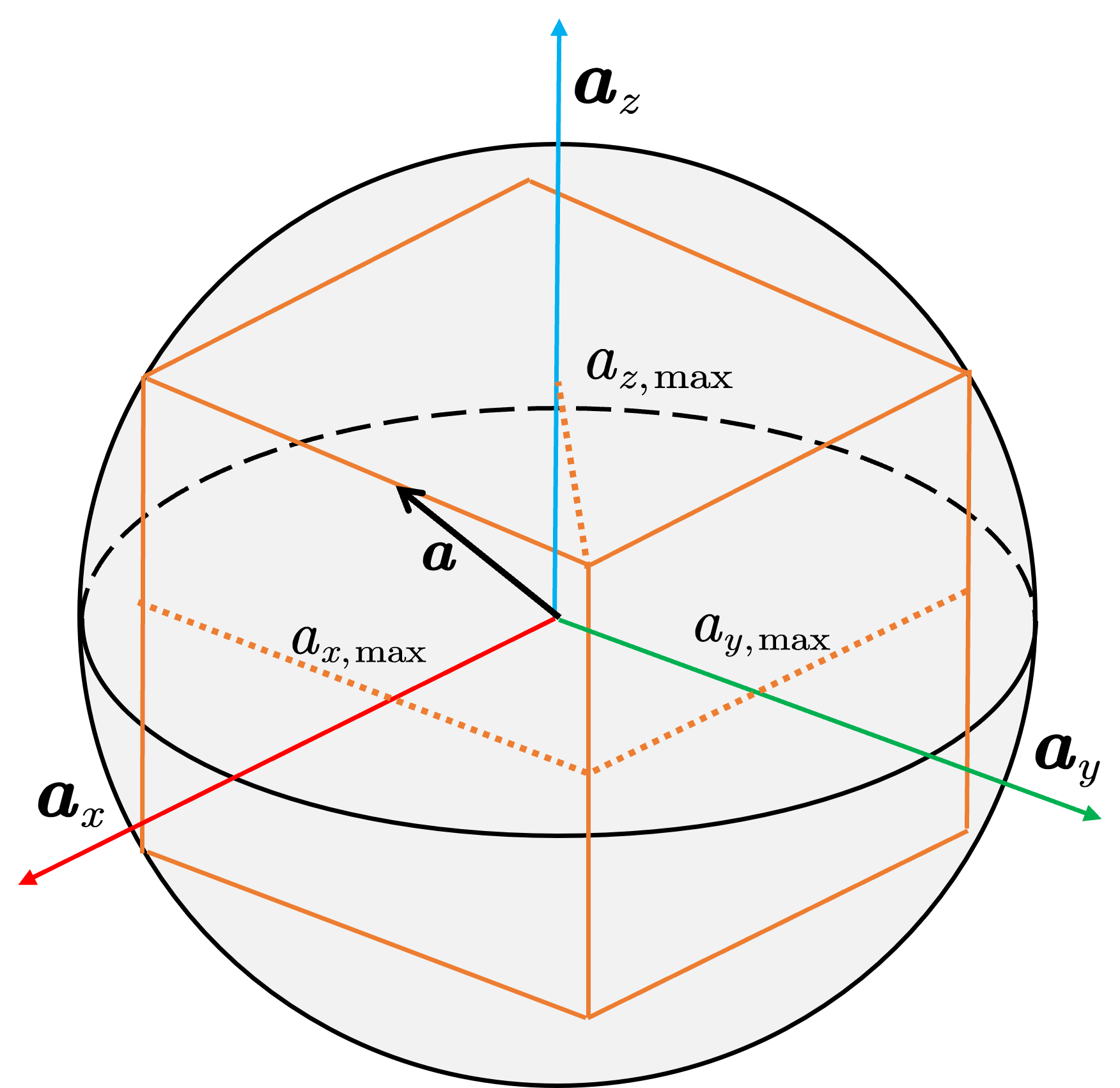}
  \label{acc_bound_fig}}
  \subfloat[]{\includegraphics[width=0.45\linewidth]{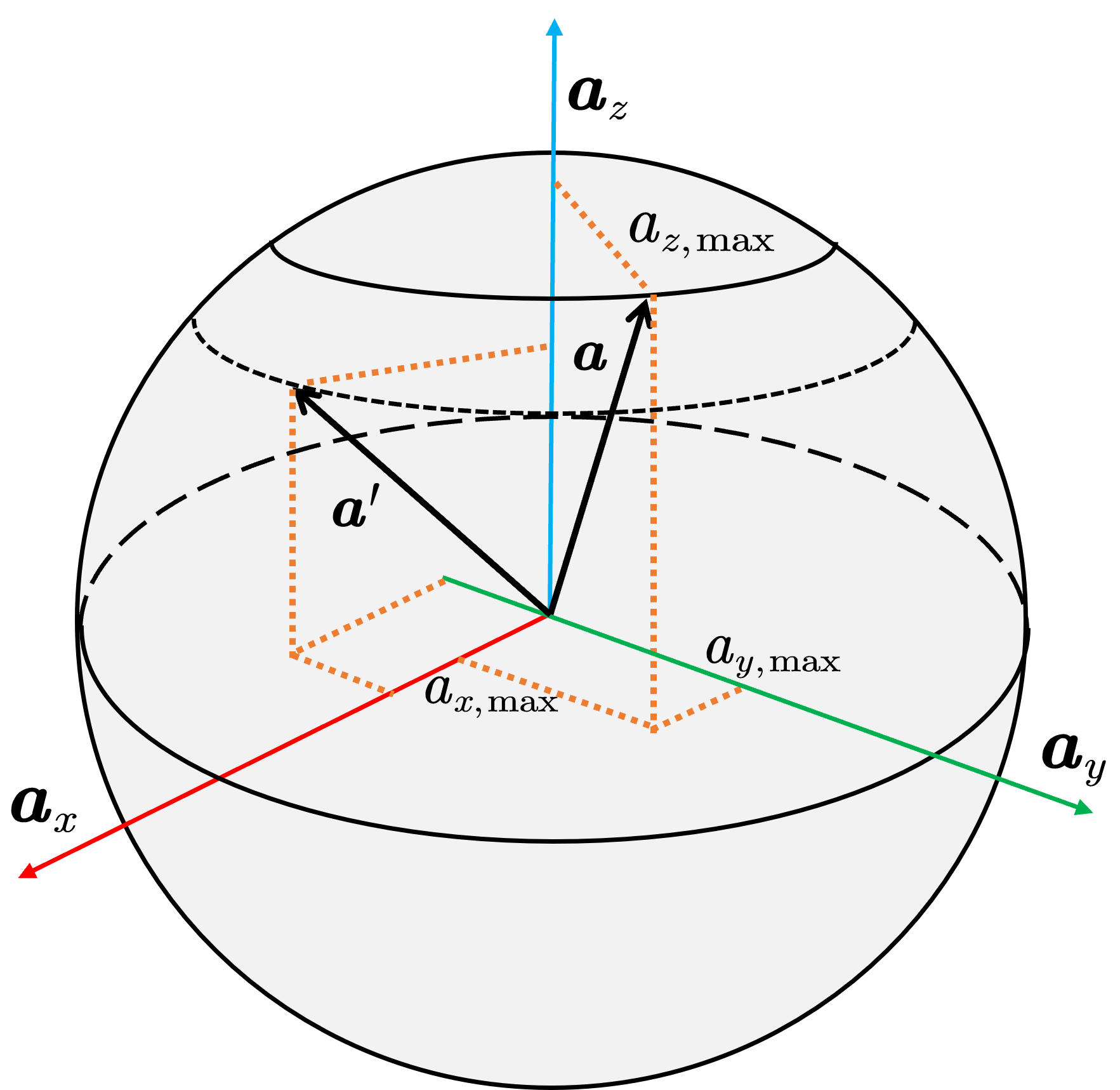}%
  \label{dyn_acc_bound_fig}}
  \caption{(a) Independent acceleration bounds limit;
           (b) Dynamic acceleration bounds limit obtained through thrust limit decomposition}
  \label{fig_acc_limit}
\end{figure}

The methods presented in the above  works can improve the time-optimal performance of the multi-rotor. 
However, the dynamic constraints imposed  by the differential flatness property or simplified dynamics with three independent acceleration bounds limits can only ensure the dynamic feasibility, but can't maximize the thrust of the multi-rotor. As illustrated  in Fig. \ref{acc_bound_fig},   the independent acceleration
bounds restrict the acceleration within a cube, preventing it from reaching the sphere determined by the thrust limit. Therefore, there is some maneuver capacity that has not been exploited.
Because of the under-actuated property, a multi-rotor has to utilize thrust in a single direction to drive movement in three directions.
Therefore, how to decompose the thrust into three directions reasonably is the key to driving multi-rotor fly to the target point fastest. 
Focusing on the problem of the time-optimal flight control of the multi-rotor, this paper proposed a guided time-optimal MPC scheme, in which the control inputs are solved by the receding-horizon optimization strategy guided by a time-optimal trajectory obtained in relaxing constraint conditions.
The primary contribution of this article lies in the thrust limit optimal decomposition method. This method can decompose the limited thrust into three directions reasonably based on the current state and the target state.  As a result, the thrust capacity of the multi-rotor can be exploited maximally.

In the follows, the time-optimal flight control problem is formulated based on the
multi-rotor dynamic model and a cascade control architecture in
Section II. Our methods of thrust limit optimal decomposition
and guided time-optimal MPC are detailed in Section III.
Experiments and results are given in Section IV. Section V
concludes this article.

\section{Problem Formulation}
\subsection{Dynamic model and control architecture}
The position of the multi-rotor is denoted by $\boldsymbol{p}$, and its orientation is described by a unit quaternion, denoted by $\boldsymbol{q}$. The rotation matrix from the body-fixed frame to the inertial frame is denoted by $\boldsymbol{R}\left(\boldsymbol{q}\right)$. The dynamics of the multi-rotor are as follows:
\begin{align}
&\dot{\boldsymbol{p}}=\boldsymbol{v}, & \dot{\boldsymbol{v}}=-\frac{f}{m}\boldsymbol{R}(q)\boldsymbol{e}_3+g\boldsymbol{e}_3   \label{eq_1}\\
&\dot{\boldsymbol{q}}=-\frac{1}{2}\left[ \begin{array}{c}
    0\\
    \boldsymbol{\omega}\\
  \end{array} \right] \otimes \boldsymbol{q}, & \dot{\boldsymbol{\omega}}=\boldsymbol{I}^{-1}(\boldsymbol{\tau }-\boldsymbol{\omega}\times (\boldsymbol{I\omega})) \label{eq_2}
\end{align}
where, $\boldsymbol{v}$ and $\boldsymbol{\omega}$ are the velocity and angular velocity, respectively. $m$ and $\boldsymbol{I}$ are the mass and inertia matrix, respectively. $g$ is the gravity acceleration.  $\boldsymbol{e}_3=\left[0, 0, 1 \right] ^T$. $f$ and $\boldsymbol{\tau }$ are the thrust and torque generated by the rotors, respectively. $\otimes$ denotes the product of two quaternions.
\begin{figure}[ht]
  \centering
  \includegraphics[width=0.9\linewidth]{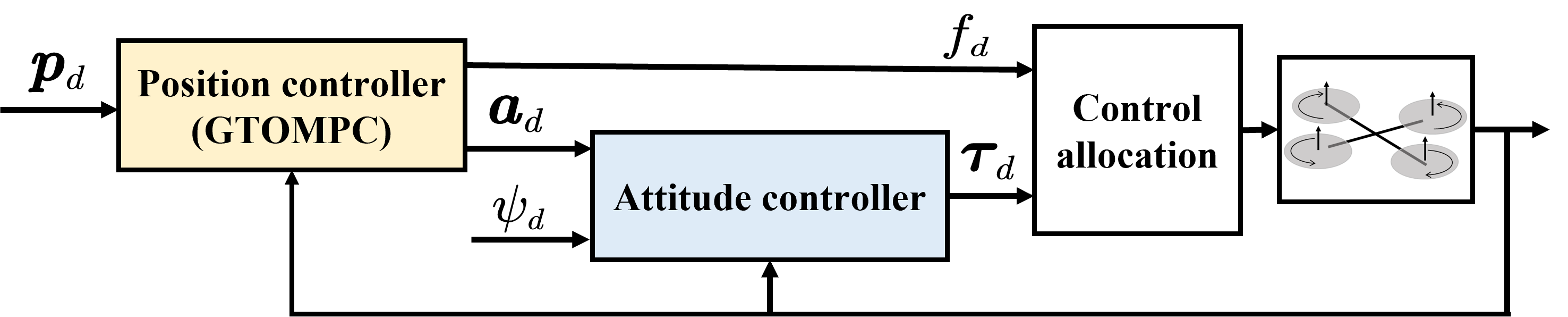}
  \caption{Control architecture}
  \label{fig_1}
\end{figure}

The cascade control architecture is used in this paper. As shown in Fig.\ref{fig_1}, the attitude controller tracks desired acceleration needed by the position controller. In this paper, only position control is taken into consideration. It is assumed that the attitude controller is fast enough, so the translational dynamics be controlled with the desired acceleration. For attitude control, methods presented in \cite{brescianini2018tilt} are used in this paper.

\subsection{Problem statement}
The translational dynamics, equation (\ref{eq_1}),  show that the movements in the three axes are coupled by $\boldsymbol{R}\left( \boldsymbol{q} \right)$.
In order to decouple the translational dynamics, firstly, the acceleration is defined as:
\begin{equation}
  \boldsymbol{a}\coloneqq -\frac{f}{m}\boldsymbol{R}\left( \boldsymbol{q} \right) \boldsymbol{e}_3+g\boldsymbol{e}_3
  \label{eq_3}
\end{equation}
The acceleration constraint can be described as follows:
\begin{equation}
 \left\| \boldsymbol{a} \right\| \le \left\|-\frac{f_{\max}}{m}\boldsymbol{R}\left( \boldsymbol{q} \right) \boldsymbol{e}_3+g\boldsymbol{e}_3\right\| \label{eq_4}
  \end{equation}
where, $f_{\max}$ is the maximum thrust of the multi-rotor. $\left\| . \right\|$ denotes the $L_2$ norm of a matrix. 
From (\ref{eq_3}), it can be observed  that the downwards maximum acceleration of $z$ axis will be larger than the upwards one due to the gravity acceleration. In this paper, to make the acceleration bound of the $z$ axis symmetrical, some maneuver capacity  is sacrificed  using a conservative acceleration constraint instead of (\ref{eq_3}). Based on the reverse triangle inequality of the $L_2$ norm,  the following inequality can be got:
\begin{equation}
  \left\|-\frac{f_{\max}}{m}\boldsymbol{R}\left( \boldsymbol{q} \right) \boldsymbol{e}_3+g\boldsymbol{e}_3\right\| \geq \left | -\left\|\frac{f_{\max}}{m}\boldsymbol{R}\left( \boldsymbol{q} \right) \boldsymbol{e}_3\right\| + \left\| g\boldsymbol{e}_3 \right\| \right |
 \end{equation}
Thus, assuming $f_{\max}\geq mg $,  the conservative acceleration constraint can be as follows:
\begin{equation}
 \left\| \boldsymbol{a} \right\| \le  \left | -\left\|\frac{f_{\max}}{m}\boldsymbol{R}\left( \boldsymbol{q} \right) \boldsymbol{e}_3\right\| + \left\| g\boldsymbol{e}_3 \right\| \right |=\frac{f_{\max}}{m}-g 
 \label{acc_conser}
\end{equation}

Combining equation (\ref{eq_1}) with (\ref{eq_3}) and (\ref{acc_conser}), 
a multi-rotor from an initial state flight to a target state in the minimum time with dynamic and thrust constraints can be formulated as the following optimal problem:
\begin{gather}
 \min_{\boldsymbol{a}\left( t \right)} \!\:\int_{0}^{t_f}dt = \min_{\boldsymbol{a}\left( t \right)} \!\: t_f \label{eq_5}\\
   s.t.  \quad  \quad  
  \left\{ \begin{matrix}\dot{\boldsymbol{p}}=\boldsymbol{v} \\ \dot{\boldsymbol{v}}=\boldsymbol{a}  \end{matrix} \right. \label{eq_6}\\
    \boldsymbol{p}\left( 0 \right) =\boldsymbol{p}_0,  \quad \boldsymbol{v}\left( 0 \right) =\boldsymbol{v}_0	\label{eq_7}\\
    \boldsymbol{p}\left( t_f \right) =\boldsymbol{p}_f, \quad \boldsymbol{v}\left( t_f \right) =\boldsymbol{v}_f   \\
 \left\| \boldsymbol{a} \right\| \le \frac{f_{\max}}{m} - g \label{eq_9} \\
    -\boldsymbol{j}_{\max}\le  \dot{\boldsymbol{a}}\le \boldsymbol{j}_{\max} \label{eq_10}
\end{gather}
where, $t_f$  is the  total time taken by the multi-rotor to reach the target state. $\boldsymbol{p}_0$ and $\boldsymbol{v}_0$ are the initial position and velocity, respectively. $\boldsymbol{p}_f$ and $\boldsymbol{v}_f$ are the target position and velocity, respectively. $\boldsymbol{j}_{\max}$ is the maximum jerk bound, which is determined by the low-level controller performance.
\section{Methodology}
In this section, the time-optimal control problem will be reformulated as a linear MPC problem using a dynamic linear acceleration constraint and a guided time-optimal cost function.

\subsection{Thrust limit optimal decomposition}
The optimal problem (\ref{eq_5})-(\ref{eq_10}) is difficult to solve since it is a free-horizon problem and the acceleration of three axes are coupled as a nonlinear constraint. As illustrated  in Fig.\ref{dyn_acc_bound_fig}, with the constraint (\ref{eq_9}), the acceleration is limited in a sphere. 
Thus, for an acceleration in any direction, its maximum bound in each axis can be obtained by decomposed thrust limit. This leads to the equivalence of the nonlinear constraint (\ref{eq_9}) to a linear constraint.
The decomposed linear acceleration constraint is as follows:
\begin{equation}
  -\boldsymbol{a}_{\max}\le \boldsymbol{a}\le \boldsymbol{a}_{\max}
  \label{eq_acc_bound}
\end{equation}
where, $\boldsymbol{a}=\left[ a_{x}, a_{y}, a_{z} \right] ^T$ and $\boldsymbol{a}_{\max}=\left[ a_{x,\max}, a_{y,\max}, a_{z,\max} \right] ^T$.

In general, the thrust limit decomposition is to decompose thrust limit as acceleration maximum bound of three axes according to the current state and the target state in every control loop, so that the nonlinear constraint will be decoupled as three dynamic linear constraints.
As a result, the time-optimal control law in the three axes decoupled case can be used directly.
In this subsection, firstly, we introduce the time-optimal control in the three axes decoupled case and the property of time-optimal control for the 3D flight task.

\subsubsection{Time-optimal control in the three axes decoupled case} 
If the flight of the multi-rotor is considered as decoupled in each axis, and the acceleration of each axis is only under maximum bound constraint, as  (\ref{eq_acc_bound}).
The time-optimal control problem in each axis can be solved by Pontryagin's Minimum Principle, and the time-optimal control law, denoted by $a_i^*(t)$,  is as follows\cite{romano2020time}:
\begin{equation}
a_{i}^{*}\left( t \right) =
  \begin{cases}
	-a_{i,\max},  &h_i >0\\
	-\mathrm{sgn}(\delta v_i) \cdot a_{i,\max},  &h_i =0\\
	a_{i,\max},  &h_i <0\\
 \end{cases}
 \label{eq_12}
\end{equation}
where, $i\in \left\{ x,y,z \right\}$. $\mathrm{sgn}() $ is the signum function. $h_i$ denotes the switch function which is defined as $h_i\left(a_{i,\max},\delta p_i,\sigma p_i,\delta v_i,\sigma v_i \right) =\sigma p_i+\frac{1}{2a_{i,\max}}\left( \sigma v_i \right) \cdot \left| \sigma v_i \right|$.
 $\delta p_i$, $\sigma p_i$, $\delta v_i$,  and $\sigma v_i$, are defined as follows:
\begin{equation}
  \begin{split}
    \delta p_i\coloneqq p_{i}-p_{i,f},\quad  & \sigma p_i\coloneqq p_{i}+p_{i,f}	\\
    \delta v_i\coloneqq v_{i}-v_{i,f},\quad  &	\sigma v_i\coloneqq v_{i}+v_{i,f} \\
  \end{split}
  \label{eq_13}
\end{equation}
The minimum time is as follows:
\begin{equation}
  t_i=\frac{1}{a_{i,\max}}\cdot T\left(a_{i,\max},\delta p_i,\sigma p_i,\delta v_i,\sigma v_i \right) 
  \label{eq_17}
\end{equation}
where, 
\begin{equation}
  \begin{split}
    & T(a_{i,\max},\delta p_i,\sigma p_i,\delta v_i,\sigma v_i)=\\
    & \begin{cases}
      \sigma v_i+\sqrt{4a_{i,\max}\delta p_i+\sigma v_{i}^{2}+\delta v_{i}^{2}},   & h_i <0 \\
      |\delta v_i|,                                                               & h_i =0 \\
      -\sigma v_i+\sqrt{-4a_{i,\max}\delta p_i+\sigma v_{i}^{2}+\delta v_{i}^{2}}, & h_i\ >0 \\
    \end{cases} \\
  \end{split}
  \label{eq_18}
\end{equation}
The time-optimal position and velocity trajectory regarding the control law (\ref{eq_12}), denoted by $p_{i}^{*}(t\!\:)$ and $v_{i}^{*}(t\!\:)$, respectively, are as follows: 
\begin{equation}
  \begin{split}
    v_{i}^{*}\left( t \right) = \int_{0}^{t}{a_{i}^{*}\left( t \right) dt},\quad   &
    p_{i}^{*}\left( t \right) = \int_{0}^{t}{v_{i}^{*}\left( t \right) dt} 
  \end{split}
  \label{eq_traj}
\end{equation}

\subsubsection{Property of time-optimal control for the 3D flight task} 
The 3D flight task is a task that needs the multi-rotor to fly in the three-axis directions. The flight times taken to arrive at the target state in the $x$, $y$, and $z$ axis direction are denoted by  $t_x$, $t_y$, and $t_z$, respectively. Thus, the 3D flight task means $t_i>0$, for $i \in \{x,y,z\}.$
For a 3D flight task, the minimum flight time of the multi-rotor using the three axes decoupled control strategy is determined by the flight time of each individual axis, and can be defined as follows:
\begin{equation}
  t_{\min}=\max \left( t_x,t_y,t_z \right)
  \label{eq_19}
\end{equation}	 	
where,  $\max \left(\right)$ is the maximum function.  
For a  given pair of initial and target states, the flight time in one axis direction is determined by the thrust decomposed in this direction. Thus, if  $t_x\ne t_y\ne t_z$,  the minimum time $t_{\min}$ can be optimized by decomposing more thrust into the direction in which the flight time is longer, until $t_x=t_y=t_z$. It can be summarized as follows:

\textbf{\textit{Property} I}: \textit{Using three axes decoupled control strategy, the multi-rotor will arrive at the target state in the three directions simultaneously under the time-optimal control law for the 3D flight task.}

The \textbf{\textit{Property} I} can be illustrated  by Fig. \ref{fig_5}.  As shown in Fig. \ref{fig_5a}, when the flight time in the three directions are equal, $t_{\min}$ is shorter than in other cases. Thus, this $t_{\min}$  can be considered as a time-optimal control solution for the 3D flight task.

\subsubsection{Thrust limit decomposition algorithm}
Based on the \textbf{\textit{Property} I}, the key to decomposing the thrust limit is to find the decomposition direction   that can make the flight time of each axis equal. We define $\boldsymbol{\widetilde{T}}=\left[ \widetilde{T}_x, \widetilde{T}_y,\widetilde{T}_z \right] ^T$ , 
where $\widetilde{T}_i = T(a_{i,\max},\delta p_i,\sigma p_i,\delta v_i,\sigma v_i)$ for $i\in \left\{ x, y, z \right\}$.

\textit{\textbf{Theorem} I: For a 3D flight task, under the control law in the three axes decoupled case, equation (\ref{eq_12}), if the flight times in each axis are equal, that is:
\begin{equation}
  t_x=t_y=t_z
  \label{eq_20}
\end{equation}
Then the direction of  $\boldsymbol{a}_{\max}$ will satisfy the following equation:
\begin{equation}
  a_{x,\max}:a_{y,\max}:a_{z,\max}=\widetilde{T}_x:\widetilde{T}_y:\widetilde{T}_z
  \label{eq_21}
\end{equation}}
The proof of \textbf{\textit{Theorem} I}:

Combing with (\ref{eq_17}) and (\ref{eq_20}), we get
\begin{equation}
  \frac{\widetilde{T}_x}{a_{x,\max}}=\frac{\widetilde{T}_y}{a_{y,\max}}=\frac{\widetilde{T}_z}{a_{z,\max}}
  \label{eq_48}
\end{equation}
which means, 
\begin{equation}
  \frac{\widetilde{T}_x}{a_{x,\max}}=\frac{\widetilde{T}_y}{a_{y,\max}}, \frac{\widetilde{T}_y}{a_{y,\max}}=\frac{\widetilde{T}_z}{a_{z,\max}}, \frac{\widetilde{T}_x}{a_{x,\max}}=\frac{\widetilde{T}_z}{a_{z,\max}}
  \label{eq_49}
  \nonumber
\end{equation}
Thus, we can get
\begin{equation}
  \frac{\widetilde{T}_x}{\widetilde{T}_y}=\frac{a_{x,\max}}{a_{y,\max}}, \frac{\widetilde{T}_y}{\widetilde{T}_z}=\frac{a_{y,\max}}{a_{z,\max}}, \frac{\widetilde{T}_x}{\widetilde{T}_z}=\frac{a_{x,\max}}{a_{z,\max}}
  \label{eq_50}
\end{equation}
That is, 
\begin{equation}
  a_{x,\max}:a_{y,\max}:a_{z,\max}=\widetilde{T}_x:\widetilde{T}_y:\widetilde{T}_z
  \label{eq_51}
\end{equation}

With \textbf{\textit{Theorem} I}, for a pair of current and target states, the optimal decomposition direction of  $a_{\max}$ can be obtained by an iterative method. It is as follows:
\begin{equation}
  a_{x,\max}^{i+1}:a_{y,\max}^{i+1}:a_{z,\max}^{i+1} =  \widetilde{T}_{x}^{i}:\widetilde{T}_{y}^{i}:\widetilde{T}_{z}^{i}
  \label{eq_22}
\end{equation}	 	
where, $i$ is the iteration index. The details are given in Alg. 1. Firstly, the direction of $\boldsymbol{a}_{\max}$  is initialized as $\left[ 1, 1, 1 \right] ^T$, then the direction of $\boldsymbol{a}_{\max}$  is iterated until the flight time of each axis are equal under tolerance $\epsilon _t$ (see Alg. 1 Line 4 - 10).  
\begin{algorithm}[ht]
  \caption{Trust limit optimal decompose}\label{alg:alg1}
  \renewcommand{\algorithmicrequire}{\textbf{Input:}}
  \renewcommand{\algorithmicensure}{\textbf{Output:}}
  \begin{algorithmic}[1]
      \Require initial state: $\boldsymbol{p}_0, \boldsymbol{v}_0$, target state: $\boldsymbol{p}_f, \boldsymbol{v}_f$
      \Ensure acceleration bound: $\boldsymbol{a}_{\max}$
      \Function{decomp\underline{~}thrust}{$\boldsymbol{p}_0, \boldsymbol{v}_0,\boldsymbol{p}_f, \boldsymbol{v}_f$}
      \State \textbf{initialize:} $ \boldsymbol{a}_{\max} \gets  \begin{bmatrix}1 & 1 & 1 \end{bmatrix}^T $
      \State \hspace{1.5cm}$ \boldsymbol{a}_{\max} \gets  \frac{\boldsymbol{a}_{\max}}{\mathbf{norm}(\boldsymbol{a}_{\max})} \cdot ({f_{\max}}/{m}-g)$
      \While  {IS\underline{~}TIME\underline{~}EQUAL($\boldsymbol{a}_{\max}, \boldsymbol{p}_0, \boldsymbol{v}_0,\boldsymbol{p}_f, \boldsymbol{v}_f$)}
      \For  {$ i = x,y,z $}
      \State $ \widetilde {T_i} \gets  T(a_{i,\max},\delta p_i,\sigma p_i,\delta v_i,\sigma v_i ) $ \Comment Equation (\ref{eq_13})-(\ref{eq_18})
      \EndFor       
      \State $ \boldsymbol{a}_{\max} \gets  \frac{\boldsymbol{{\widetilde T}}}{\mathbf{norm}(\boldsymbol{\widetilde T})} \cdot ({f_{\max}}/{m}-g)$   \Comment Equation (\ref{eq_22})
      \EndWhile
      \State \textbf{return} $\boldsymbol{a}_{\max}$
      \EndFunction
      \State
      \Function{is\underline{~}time\underline{~}equal }{$\boldsymbol{a}_{\max}, \boldsymbol{p}_0, \boldsymbol{v}_0,\boldsymbol{p}_f, \boldsymbol{v}_f$}
      \For {$ i = x,y,z $}
      \State $ {t_i} \gets \frac{1}{{a_{i,\max}}} \cdot T(a_{i,\max}, \delta p_i,\sigma p_i,\delta v_i,\sigma v_i ) $ \Comment Equation (\ref{eq_13})-(\ref{eq_18})
      \EndFor
      \For {$ i,j = x,y,z $}
      \If{$t_i > \epsilon _t$, $t_j > \epsilon _t$ and $|t_i-t_j| > \epsilon _t$}
      \State \textbf{return} False
      \EndIf
      \EndFor
      \State \textbf{return} True
      \EndFunction
  \end{algorithmic}
\end{algorithm}

\subsection{Guided time-optimal MPC}
In (\ref{eq_12}),   $a^*\left(t\right)$  switches between the upper and lower bound, behaving like a step signal. Thus, it is not safe enough to use  $a^*\left(t\right)$  directly as the desired acceleration to be tracked by the attitude controller.   
However, it can be seen as a solution to the problem (\ref{eq_5})-(\ref{eq_10}) in the relaxing constraint conditions that ignore jerk constraint. Thus, the regarding trajectory $\boldsymbol{p}^*\left(t\right)$, equation (\ref{eq_traj}), can be used as a time-optimal guidance trajectory.  We define the guided time-optimal cost function as follows:
\begin{equation}
  J=\int_0^{t_{\min}}{\left(\rho _0\left\| \boldsymbol{p}(t\!\:)-\boldsymbol{p}^*(t\!\:) \right\| ^2+\rho _1\left\| \boldsymbol{v}(t\!\:)-\boldsymbol{v}^*(t\!\:) \right\| ^2 \right)}dt
  \label{eq_23}
\end{equation}
where, $\rho _0$ and $\rho _1$ are the weight of position and velocity, respectively. 
Replacing  $\int_{0}^{t_f}{dt}$ by $J$,  the optimal problem (\ref{eq_5})-(\ref{eq_10})  can be converted into a quadratic optimal control problem in the finite time horizon. 

 After the thrust limit decomposition and combining with the guided time-optimal cost function, problem (\ref{eq_5})-(\ref{eq_10}) can be converted into a linear MPC problem. By discretizing it with a time step size of $\Delta t$, we can get:
\begin{equation}
  \min_{\{a[0],...,a[N_p]\}} \sum_{k=0}^{N_p}{\left(\rho _0\left\| \boldsymbol{p}[k]-\boldsymbol{p}^*[k] \right\| ^2+\rho _1\left\| \boldsymbol{v}[k]-\boldsymbol{v}^*[k] \right\| ^2 \right)}
  \label{eq_24}
\end{equation}
\begin{align}
    \left[ \begin{array}{c}
    \boldsymbol{p}\left[ k+1 \right]\\
    \boldsymbol{v}\left[ k+1 \right]\\
  \end{array} \right] &=\boldsymbol{A}_d\left[ \begin{array}{c}
    \boldsymbol{p}\left[ k \right]\\
    \boldsymbol{v}\left[ k \right]\\
  \end{array} \right] +\boldsymbol{B}_d\boldsymbol{a}\left[ k+1 \right] 
  \label{eq_25}\\
   &\boldsymbol{p}\left[ 0 \right] =\boldsymbol{p}_0, \boldsymbol{v}\left[ 0 \right] =\boldsymbol{v}_0
  \label{eq_26}
\end{align}
for  $k\ge 0$,
\begin{align}
    & -\boldsymbol{a}_{\max}<\boldsymbol{a}\left[ k \right] < \boldsymbol{a}_{\max}\\
   -\boldsymbol{j}_{\max} &\varDelta t<\boldsymbol{a}\left[ k+1 \right] -\boldsymbol{a}\left[ k \right] < \boldsymbol{j}_{\max}\varDelta t \label{eq_27}
\end{align}
where,  $\boldsymbol{p}\left[ k \right]$ and $\boldsymbol{v}\left[ k \right]$ denote the position and velocity of the multi-rotor at $k$ time step, respectively.  
$\boldsymbol{A}_d=\left[ \begin{matrix}
	\boldsymbol{I}_{3\times 3}&		\varDelta t\cdot \boldsymbol{I}_{3\times 3}\\
	\boldsymbol{O}_{3\times 3}&		\boldsymbol{I}_{3\times 3}\\
\end{matrix} \right]$,  
$\boldsymbol{B}_d=\left[ \begin{array}{c}
	\frac{1}{2}\varDelta t^2\cdot \boldsymbol{I}_{3\times 3}\\
	\varDelta t\cdot \boldsymbol{I}_{3\times 3}\\
\end{array} \right]$.
$a_\text{max}$ is the acceleration bound limit decomposed by the Alg. 1.  

This control scheme (containing thrust limit optimal decomposition, guided time-optimal cost function, and receding-horizon optimization strategy) is called GTOMPC (Guided Time-optimal MPC), 
which means that the control law is solved under the guidance of a time-optimal trajectory in the relaxing
constraint condition.

\section{Experiments and results}
Several experiments have been conducted to validate methods proposed in Section III, and the experiments and results will be introduced in detail in this section.

\subsection{Thrust limit decomposition experiments}
\begin{figure}[ht]
  \centering
  \subfloat[]{\includegraphics[width=0.5\linewidth]{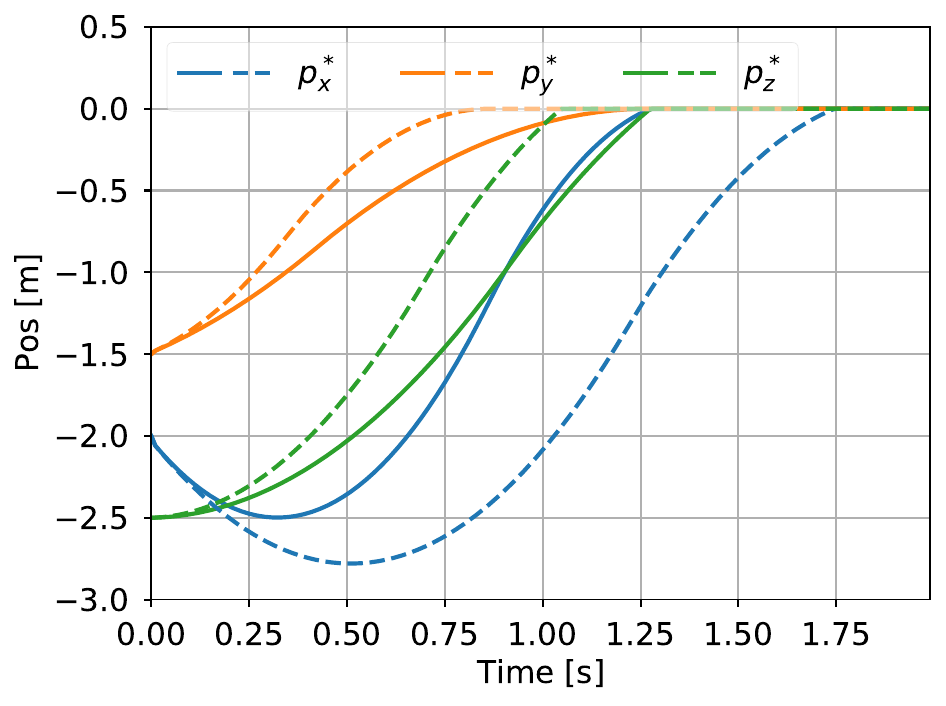}
  \label{fig_5a}}
  \subfloat[]{\includegraphics[width=0.5\linewidth]{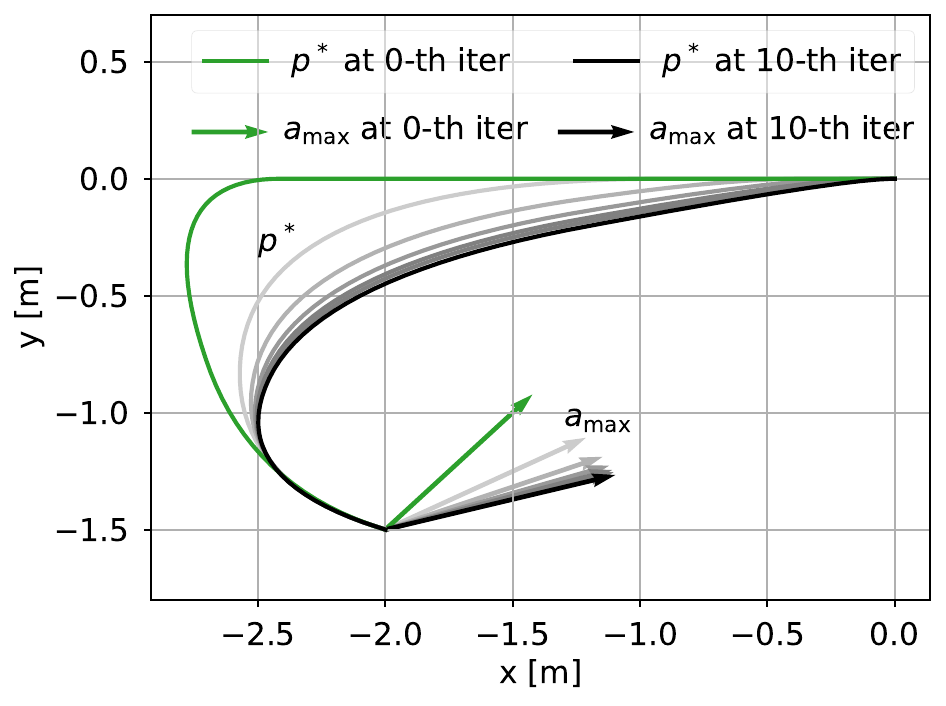}%
  \label{fig_5b}}
  \caption{(a)  $\boldsymbol{p}^*$  regarding to different $a_{\max}$  
                           (dashed line regarding to the initial $a_{\max}=[5.774, 5.774, 5.774]^T (m/s^2)$ ; solid line  regarding to the decomposed $a_{\max}=[9.028, 2.225, 3.612]^T (m/s^2)$ );  
                           (b)  convergence diagram  of the direction of $a_{\max}$  and regarding  $\boldsymbol{p}^*$  as the number of decomposition iteration increases (demonstrated in the $x$-$o$-$y$ plane; grey lines regarding 1-th $\sim$ 9-th iteration).}
  \label{fig_5}
\end{figure}

To validate the convergence and the iteration efficiency  of the thrust limit optimal decomposition method (Alg. 1), 
1000 pairs of random initial and target states were tested in the thrust limit decomposition experiments.
 For a specific pair of data (initial state: $p_0=[-2, -1.5, -2.5]^T (m)$ and $v_0=[-3, 1, 0]^T (m/s)$; target state: $p_f=[0, 0, 0]^T (m)$ and $v_f=[1, 0, 2]^T [(m/s)$; thrust limit: ${f_{\max}}/{m}-g=10 (m/s^2)$), the acceleration maximum bound $a_{\max}$ is converged from $[5.774, 5.774, 5.774]^T(m/s^2)$ to $[9.028, 2.225, 3.612]^T(m/s^2)$ using Alg. 1 .
The  time-optimal guidance trajectory $\boldsymbol{p}^*\left(t\right)$, (\ref{eq_traj}),  regarding  the initial and decomposed $a_{\max}$   are shown in Fig.\ref{fig_5}. 
As shown in Fig.\ref{fig_5a},  after thrust limit  decomposition, with more thrust limit in the  $x$ and $y$ axis, the flight time in the three directions are equal. As a result, the minimum time $t_{\min}$, defined in (\ref{eq_19}), is improved from $1.75s$ to $1.25s$.  

\begin{figure}[ht]
  \centering
  \includegraphics[width=0.9\linewidth]{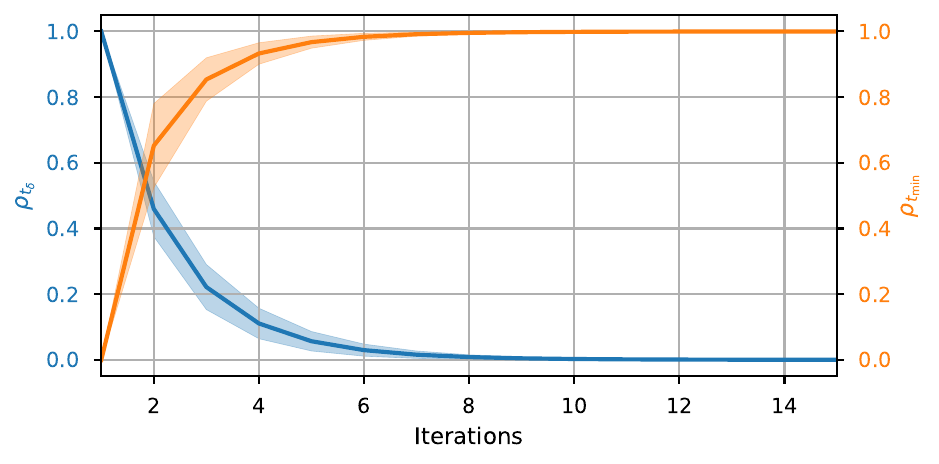}
  \caption{The mean and standard deviation of $\rho _{t_{\delta}}$ and $\rho _{t_{\min}}$ of the thrust limit decomposition obtained over 1000 pairs of random data.}
  \label{fig_decomp_mean_iter}
\end{figure}

To quantify the iteration efficiency of Alg. 1, two indicators are defined, as follows:
\begin{align}
    &\rho _{t_{\delta}}^{i}={{t_{\delta}^{i}}/{t_{\delta}^{0}}}\\
    &\rho _{t_{\min}}^{i}={\left( t_{\min}^{0}-t_{\min}^{i}\right)}/{\left(t_{\min}^{0}-t_{\min}^* \right)}
\end{align}
where, 
$t_{\delta}^{i}=\max \left( \left| t_{x}^{i}-t_{y}^{i} \right|,\left| t_{y}^{i}-t_{z}^{i} \right|,\left| t_{x}^{i}-t_{z}^{i} \right| \right)$; 
$t_{\min}^*$ is the final convergence value of minimum time after iterated, defined by (\ref{eq_19});
$\rho _{t_{\delta}}^{i}$  quantifies the ratio of differential arrival times among the three directions at the $i$-th iteration, normalized by $t_{\delta}^{0}$.
Thus, the $\rho _{t_{\delta}}^{i} = 0$ when the multi-rotor can arrive at the target state in the three directions simultaneously. 
$\rho _{t_{\min}}^{i}$ quantifies the optimization ratio of the minimum time at $i$-th iteration.
Thus, the $\rho _{t_{\min}}^{i} = 1$ when the multi-rotor can arrive at the target state in the shortest time. 
As the number of decomposition iteration increases, the mean and standard deviation of $\rho _{t_{\delta}}$ and $\rho _{t_{\min}}$ of the test dates are shown in Fig.\ref{fig_decomp_mean_iter}.
As shown in  Fig.\ref{fig_decomp_mean_iter}, for any pair of initial and target states, the $\rho _{t_{\delta}}$ converges to $0$ and $\rho _{t_{\min}}$ converges to $1$ at about 10-th iteration.

\subsection{Flight experiments}
The platform used in the flight experiments is a hex-rotor, and its total mass and arm length are $2.4kg$ and $0.275m$, respectively.
The hex-rotor is equipped with a PX4FMU flight computer, in which the attitude controller and state estimator are running.
The onboard computer used to run the GTOMPC controller is NVIDIA Jetson NX. 
In the onboard computer, the ROS is used as a communication software framework for different application nodes. 
The QP solver used in the GTOMPC controller is OSQP $\footnote{https://osqp.org}$. 
The experiments have been conducted outdoors.
\begin{figure}[ht]
  \centering
  \subfloat[]{\includegraphics[width=0.8\linewidth]{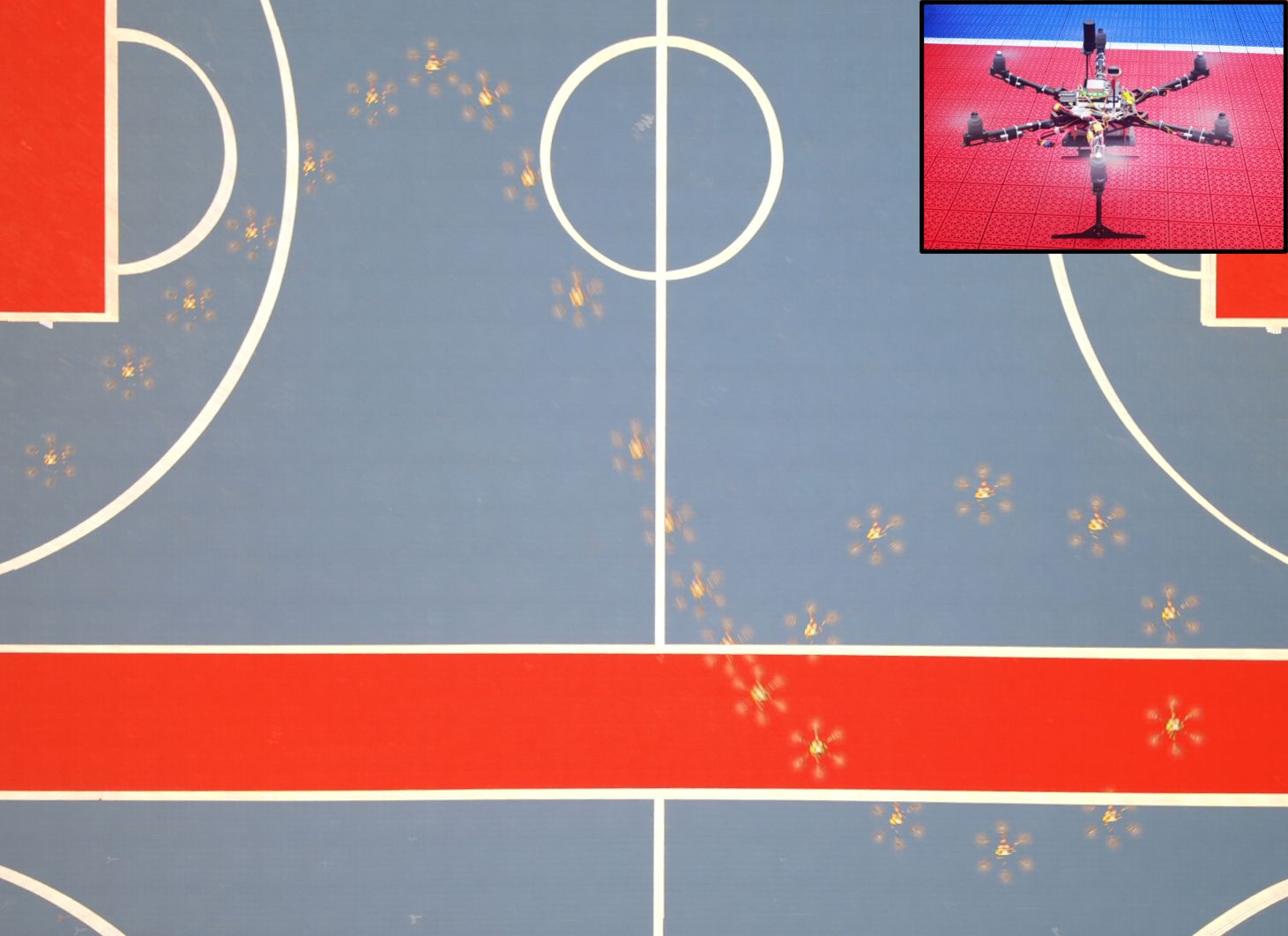}\label{fig_path_a}}
  \hfil  
  \subfloat[]{\includegraphics[width=0.9\linewidth]{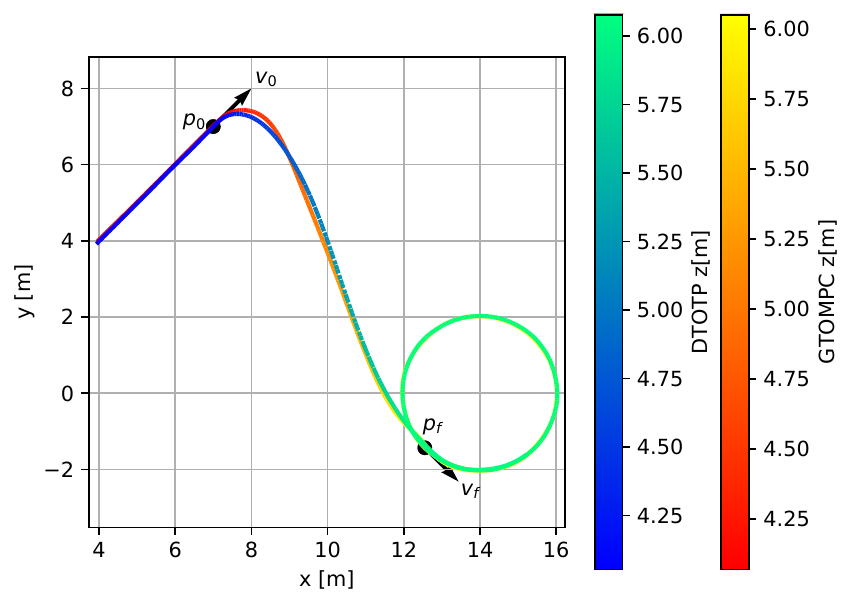}\label{fig_path_b}} 
  \caption{The hex-rotor is controlled by the proposed method to fly from a line trajectory to switch into a circular trajectory.
  }
  \label{fig_path}
\end{figure}

We design a no-reset to no-reset flight task to validate that the proposed methods can control the multi-rotor flight with time-optimal performance.
The no-reset to no-reset flight task is to control the hex-rotor flight from a line trajectory switch into a circular trajectory, as shown in Fig.\ref{fig_path}.
The line trajectory starts at position $[4, 4, 4]^T (m)$ and ends at position $[7, 7, 4]^T (m)$, and with a constant velocity $[1, 1, 0]^T(m/s)$.
The circular trajectory is: 
$
    \boldsymbol{p}(t)=\left[
    14 +  2\cos(\frac{1}{5}\pi t - \frac{3}{4}\pi),
    2\sin(\frac{1}{5}\pi t- \frac{3}{4}\pi),
    6
\right]^T (m)
$.
Thus, the start and final state of the no-reset to no-reset flight task are: $\boldsymbol{p}_0=[7, 7, 4]^T(m)$, $\boldsymbol{v}_0=[1, 1, 0]^T (m/s)$ and
$\boldsymbol{p}_f=[12.516, -1.446, 6]^T(m)$, $\boldsymbol{v}_f=[0.888, -0.888, 0]^T(m/s)$, respectively. The main parameters of the GTOMPC controller are listed in Table \ref{tab_1}. 

\begin{table}[ht]
  \begin{center}
    \caption{Params of GTOMPC controller}
    \label{tab_1}
    \begin{tabular}{ c  c  c }
      \hline
      Symbol                & Description	                  & Value\\        
      \hline
      ${f_{\max}}/{m}-g$          & Thrust limit                  & $5m/s^2$\\
      $\rho _0$             & Zero-order proximity weight	  & $1$\\
      $\rho _1$             & First-order proximity weight  & $0.5$\\
      $\Delta t$            & Discretized time step size    & $0.05s$ \\
      $N_{p}$          & Prediction step       & $30$\\
      $j_{\max}$            & Maximum jerk                  & $5m/s^3$\\
      \hline
    \end{tabular}
  \end{center}
\end{table}

\begin{figure}[ht]
  \centering
  \subfloat[]{\includegraphics[width=0.8\linewidth]{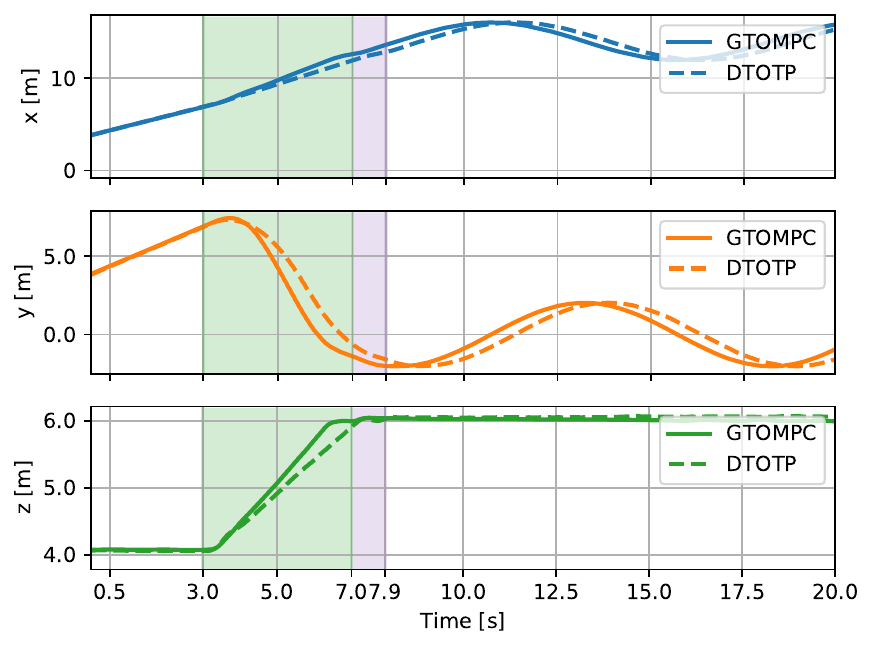}
  \label{fig_7a}
  }
  \hfil  \subfloat[]{\includegraphics[width=0.8\linewidth]{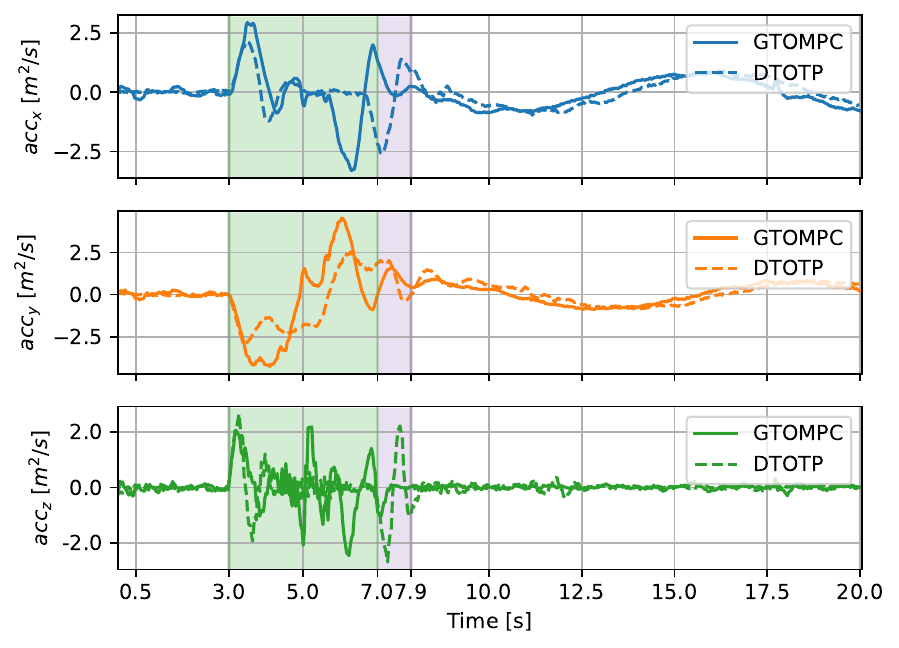}
  \label{fig_7b}
  }
  \caption{Flight performance comparison in the no-reset to no-reset task (the GTOMPC completed the task in $t=3.0s$-$7.0s$
  and the DTOTP completed the in $t=3.0s$-$7.9s$): 
                                          (a) position;
                                          (b) desired acceleration.}
  \label{fig_7}
\end{figure}

In the flight experiments, the proposed GTOMPC method is compared with decoupled time-optimal trajectory planning (DTOTP) method in which the dynamics are decoupled directly and the acceleration maximum of each axis has the same constant value\textsuperscript{\cite{hehn2015real}}.
The experimental results are shown in Fig.\ref{fig_7}. 

As shown in Fig.\ref{fig_7a}, the hex-rotor using the proposed GTOMPC method can complete the no-reset to no-reset task with about $4s$, which is faster 
than it using DTOTP method (about $4.9s$). 
As shown in Fig.\ref{fig_7b},  with the thrust limit decomposition, more thrust is decomposed into the $x$ and $y$ axis directions 
which is the reason why the GTOMPC has better time-optimal performance than DTOTP. 
The GTOMPC is a linear MPC and the thrust decomposition can be completed at about 10-th iteration, so the GTOMPC can be solved efficiently. 

\section{Conclusion}
Focusing on the time-optimal problem of the multi-rotor, this paper proposed a thrust limit optimal decomposition method and a GTOMPC control scheme.
The thrust limit optimal decomposition method can decouple the dynamics by decomposing the thrust limit into three directions, reasonably.
The GTOMPC can convert the time-optimal control problem into a linear MPC problem with a time-optimal guidance trajectory, which is got after the thrust limit decomposed.
Experimental results show that the proposed method can decompose the thrust limit efficiently, and with the GTOMPC control scheme the hex-rotor can complete the no-reset to no-reset task with better time-optimal performance 
than the decoupled time-optimal trajectory planning method.



\bibliographystyle{IEEEtran}
\bibliography{IEEEabrv,bib/LCSS}\ 


\end{document}